\documentclass[letterpaper]{article}

\PassOptionsToPackage{numbers, compress}{natbib}
\usepackage{fullpage}

\usepackage{times}
\usepackage{soul}
\usepackage{url}
\usepackage[hidelinks]{hyperref}
\usepackage[utf8]{inputenc}
\usepackage[small]{caption}
\usepackage{graphicx}
\usepackage{amsmath}
\usepackage{booktabs}
\usepackage{algorithm}
\usepackage{algorithmic}
\usepackage{wrapfig} 
\usepackage{xcolor}
\usepackage{soul}
\usepackage{amsmath,amsfonts,amsthm,amssymb,amsbsy,mathrsfs}
\usepackage[utf8]{inputenc}
\usepackage{helvet}  
\usepackage{courier}  
\usepackage{url}  
\usepackage{graphicx}  
\usepackage{hyperref}       
\usepackage{url}
\usepackage{booktabs}       
\usepackage{nicefrac}       
\usepackage{microtype}      
\newtheorem{thm}{Theorem}

\newtheorem{assumption}[thm]{Assumption}

\usepackage{bm}
\usepackage{graphicx, xcolor}
\usepackage{bchart}
\usepackage{pgfplots}
\usepackage{algorithm}
\usepackage{algorithmic}
\usepackage[english]{babel}
\usepackage{comment}
\usepackage{extarrows}
\usepackage{multirow}

\usepackage{booktabs}
\usepackage{subfigure}
\usepackage{bm}
\usepackage{algorithm}
\usepackage{algorithmic}
\usepackage{paralist}
\usepackage{hhline}
\def \S {\mathbf{S}}
\def \A {\mathbf{A}}
\def \R {\mathbb{R}}
\def \bL {\mathbf{L}}
\def \N {\mathbf{N}}

\def \E {\mathbf{E}}
\def \T {\bm{T}}
\def \DTW {\mathrm{DTW}}
\def \X {\mathbf{X}}
\def \x {\mathbf{x}}
\def \Q {\mathbf{Q}}
\def \x {\mathbf{x}}
\def \e {\mathbf{e}}
\def \Ocal {\mathcal{O}}

\newcommand{\argmin}{\arg\!\min}

\makeatother
\usepackage[utf8]{inputenc} 
\usepackage[T1]{fontenc}    
\usepackage{hyperref}       
\usepackage{url}            
\usepackage{booktabs}       
\usepackage{amsfonts}       
\usepackage{nicefrac}       
\usepackage{microtype}      
\usepackage{authblk}

\title{\textup{Similarity Preserving Representation Learning for\\ Time Series Clustering}}
\author[$\star$]{Qi Lei}
\author[$\mathsection$]{\ \ Jinfeng Yi}
\author[$\dagger$]{\ \ Roman Vaculin}
\author[$\dagger$]{\ \ Lingfei Wu}
\author[$\star\ddagger$]{\ Inderjit S. Dhillon}
\affil[ ]{$^\star$ UT Austin \ \  $^\mathsection$ JD AI Research \ \   $^\dagger$ IBM Research \ \  $^\ddagger$ Amazon}
\affil[ ]{leiqi@oden.utexas.edu\ \ \ \   yijinfeng@jd.com}

\begin{document}

\maketitle

\begin{abstract}
A considerable amount of clustering algorithms take instance-feature matrices as their inputs. As such, they cannot directly analyze time series data due to its temporal nature, usually unequal lengths, and complex properties. This is a great pity since many of these algorithms are effective, robust, efficient, and easy to use. In this paper, we bridge this gap by proposing an efficient representation learning framework that is able to convert a set of time series with various lengths to an instance-feature matrix. In particular, we guarantee that the pairwise similarities between time series are well preserved after the transformation, thus the learned feature representation is particularly suitable for the time series clustering task. 
Given a set of $n$ time series, we first construct an $n\times n$ partially-observed similarity matrix by randomly sampling $\mathcal{O}(n \log n)$ pairs of time series and computing their pairwise similarities. We then propose an efficient algorithm that solves a non-convex and NP-hard problem to learn new features based on the partially-observed similarity matrix. By conducting extensive empirical studies, we show that the proposed framework is more effective, efficient, and flexible,  compared to other state-of-the-art time series clustering methods.
\end{abstract}
\vspace{-3pt}
\section{Introduction} 
\vspace{-2pt}Modeling time series data is important but highly challenging. It is considered by \cite{yang200610} as one of the $10$ most challenging problems in data mining. Although time series clustering has attracted increasing attention, the time series clustering algorithms are still much fewer than the clustering algorithms developed for static data. The latter category of models, which usually take instance-feature matrices as their inputs, cannot directly partition time series data due to its temporal nature, typically unequal lengths, and complex properties~\cite{langkvist2014review}. This is a great pity since many static clustering algorithms are effective, robust, efficient, and easy to use. Introducing them to time series clustering can significantly advance this field.

In this work, we bridge this gap 
by proposing an efficient 
unsupervised representation learning framework that can convert a set of uni- or multi-dimensional time series data with equal or unequal lengths to an instance-feature matrix. In particular, the learned features preserve the pairwise similarities between the raw time series data, thus are particularly suitable to the time series clustering problem. 
Notably, the proposed framework is flexible to any time series distance or similarity measures such as Mikowski distance, cross-correlation, Kullback-Leibler divergence, dynamic time warping (DTW) similarity, move-split-merge (MSM) distance, and short time series (STS) distance. Here, we slightly abuse the notation and use the term similarity measure to denote both similarity and distance measures, as they are usually interchangeable. In this paper, we use widely-used DTW similarity as an example to illustrate our approach, and then empirically show that the proposed approach also works well with other time series similarity measures.

Given a total of $n$ uni- or multi-dimensional time series, our first step generates an $n\times n$ similarity matrix $\A$ with $\A_{ij}$ equaling to the DTW similarity between the time series $i$ and $j$. However, computing all the pairwise similarities requires calling the DTW $\mathcal{O}(n^2)$ times, which is time-consuming when $n$ is large. As a concrete example, generating a full similarity matrix for $n=150,000$ time series of length $30$ takes more than $28$ hours on an Intel Xeon $2.40$ GHz processor with $256$ GB of main memory. To significantly reduce the running time, we note that time series similarity measures, including DTW, usually capture the co-movements of time series, which has shown to be driven by only a small number of latent factors~\cite{stock2005implications}. This suggests that 
the similarity matrix $\A$ can be well approximated by a low-rank matrix.
According to the theory of (noisy) matrix completion~\cite{lafond2015low,sun2015guaranteed}, only $\mathcal{O}(n\log n)$ randomly sampled entries are needed to recover an $n \times n$ low-rank matrix. This allows us to sample only  $\mathcal{O}(n\log n) $ pairs of time series to generate a partially-observed similarity matrix $\tilde \A$. In this way, the time spent on generating similarity matrix is significantly reduced by a factor of $\mathcal{O}(n/\log n)$. For $n=150,000$ time series of length $30$, it only takes 3 minutes to construct a partially observed similarity matrix with $[20 n \log n]$ observed entries. 


Given the generated partially-observed similarity matrix $\tilde \A$, our second step learns a new feature representation for $n$ time series such that their pairwise DTW similarities can be well approximated by the inner products of new features. To this end, we solve a symmetric matrix factorization problem to factorize $\tilde \A$,
i.e., learn a $d$-dimensional matrix $\X\in \R^{n\times d}$ such that $P_{\Omega}{(\tilde \A)}\approx P_{\Omega}{(\X\X^\top)}$, where $P_{\Omega}$ is a matrix projection defined on the observed set $\Omega$. Despite its relatively simple formulation, this optimization problem is NP-hard and non-convex. To address this challenge, we propose a highly efficient and parameter-free exact cyclic coordinate descent algorithm. By wisely updating variables with the sparsely observed entries in $\tilde \A$, the proposed algorithm incurs a very low computational cost, and thus can learn new feature representations in an extremely efficient way. For example, with $n=150,000$ and $d=15$, it only takes the proposed algorithm $25$ seconds to learn new features.
We summarize our contributions as below:
\begin{enumerate}
    \item We bridge the gap between time series data and static clustering algorithms by learning a feature representation that preserves the pairwise similarities of the raw time series data. The underlying low-rank assumption of the similarity matrix is verified both theoretically and empirically.  
    \item We propose a parameter-free algorithm for symmetric matrix factorization on a partially-observed matrix. 
    \item We conduct extensive experiments on over $80$ real-world time series datasets. The results show that our learned features working with some simple static clustering methods like $k$Means can  significantly outperform the state-of-the-art time series clustering algorithms in both accuracy and efficiency. 
\end{enumerate}

\vspace{-3pt}
\section{Related Work}
\vspace{-2pt}
In this section, we briefly review the existing work on learning feature representations for time series data. 
A majority of them were designed for time series classification instead of clustering. 
A family of methods uses a set of derived features to represent time series. For instance, \cite{nanopoulos2001feature} proposed to use the mean, standard deviation, kurtosis, and skewness of time series to represent control chart patterns. The authors in
\cite{wang2006characteristic} introduced a set of features such as trend, seasonality, serial correlation, chaos, nonlinearity, and self-similarity to partition different types of time series. \cite{deng2013time} used some easy to compute features such as mean, standard deviation and slope temporal importance curves to guide time series classification. To automate the selection of features for time series classification, the authors in \cite{fulcher2014highly} proposed a greedy forward method that can automatically select features from thousands of choices. Shapelet-based classifiers~\cite{DBLP:journals/datamine/HillsLBMB14} extract shapelet features from the time series, and then convert the time series to a regular feature table.~\cite{DBLP:journals/datamine/Kate16} proposed to use the DTW distances from a training example to create new features for time series data. Besides, several techniques have been proposed to represent time series by a certain types of transformation, such as discrete Fourier transformation~\cite{faloutsos1994fast}, discrete cosine transformation~\cite{korn1997efficiently}, Laplace transformation~\cite{narita2007learning,hayashi2005embedding}, discrete wavelet transformation~\cite{chan1999efficient}, 
piecewise aggregate approximation~\cite{keogh2001dimensionality}, and symbolic aggregate approximation~\cite{lin2007experiencing}. 
Also, deep learning models such as Elman recurrent neural network~\cite{elman1990finding} and long short-term memory~\cite{hochreiter1997long} are capable of modeling complex structures of time series data and learn a layer of feature representations. 

Despite the remarkable progress, most feature representations learned by these algorithms are problem-specific and are not general enough for applications in multiple domains. Besides, these learned features cannot preserve similarities of the raw time series data, thus are not suitable for the clustering problem that is sensitive to the data similarity. 
These limitations inspire us to propose a problem-independent and similarity preserving representation learning framework for time series clustering.
\vspace{-3pt}
\section{Similarity Preserving Representation Learning for Time Series Clustering} \vspace{-2pt}
In this section, we first present the general framework of our similarity preserving time series representation learning method, and then propose an extremely efficient algorithm that is significantly faster than a naive implementation.
\subsection{Problem Definition and General Framework}
Given a set of $n$ time series $\mathcal{T}=\{T_1, \cdots, T_n\}$ with equal or unequal lengths, our goal is to convert them to a matrix $\X\in \R^{n\times d}$ such that the time series similarities are well preserved after the transformation.
Specifically, we aim to learn a mapping function $f: T\rightarrow \R^d$ that satisfies
\begin{eqnarray}
\mathrm{S}(T_i, T_j)\approx \langle f(T_i),f(T_j)\rangle\ \  \forall{i,j}\in [n],\label{eq:1}
\end{eqnarray}
where $\langle \cdot , \cdot\rangle$ stands for the inner product, 
a common similarity measure in analyzing static data. $\mathrm{S}(\cdot, \cdot)$ denotes the pairwise time series similarity that can be computed by a number of functions. In this work, we use dynamic time warping (DTW) algorithm as an example to illustrate our approach. 
By warping sequences non-linearly in the time dimension, DTW can calculate an optimal match between two given temporal sequences with equal or unequal lengths. Due to its superior performance, DTW has been successfully applied to a variety of applications, including computer animation~\cite{muller2007dtw}, surveillance~\cite{sempena2011human}, gesture recognition~\cite{celebi2013gesture}, signature matching~\cite{efrat2007curve}, 
and speech recognition~\cite{muda2010voice}.

Normally, DTW outputs a pairwise distance between two temporal sequences, thus we need to convert it to a similarity score. Since the inner product space can be induced from the normed vector space using
$\langle x,y\rangle 
=(\|x\|^2+\|y\|^2-\|x-y\|^2)/2$~\cite{adams2004knot}, we generate the DTW similarity by
\begin{eqnarray}
\mathrm{S}(T_i,T_j)\!=\!\frac{\DTW(T_i,\emph{0})^2\!\!+\!\DTW(T_j,\emph{0})^2\!\!-\!\DTW(T_i,T_j)^2}{2},\label{eq:11}
\end{eqnarray}
where $\emph{0}$ denotes the length one time series with entry $0$. The similarity computed via the above equation is a more numerically-stable choice than some other similarity measures such as the reciprocal of distance. This is because when two time series are almost identical, their DTW distance is close to $0$ and thus its reciprocal tends to infinity. An alternative way to convert DTW is via the Laplacian transformation  $e^{-\text{DTW}^2(T_i,T_j)/t}$~\cite{hayashi2005embedding}, where $t$ is a hyperparameter. However, our experimental results shows that it yields a worse performance than the proposed method (\ref{eq:11}).



In order to learn the matrix $\X$, an intuitive idea is to factorize the similarity matrix $\A\in \R^{n\times n}$ where $\A_{ij}=\mathrm{S}(T_i,T_j)$.
In more detail, this idea consists of two steps, i.e., a similarity matrix construction step and a symmetric matrix factorization step. In the first step, we construct $\A$ by calling the DTW oracle at least $(n+1)n/2$ times to compute the pairwise similarites and the distance.
In the second step, we learn an optimal data-feature matrix $\X$ by solving the following optimization problem
\begin{equation}
\min_{\X\in \R^{n\times d}}\ \ \ \|\A-\X\X^\top\|_F^2,\label{eq:4}
\end{equation}
where $i$-th row of $\X$ indicates the feature vector of sample $i$.
The problem \eqref{eq:4} has a closed form solution, i.e.,
\[\X=\Q_{1:n,1:d}\times \sqrt{\bm\Lambda_{1:d,1:d}}\ ,\]
where $\A=\Q\bm\Lambda\Q^\top$ and the notation $\Q_{1:k,1:r}$ represents the upper left $k$ by $r$ sub-matrix of $\Q$.

Although the inner products of features in $\X$ well preserve the DTW similarities of the raw time series, the idea described above is impractical since both construction and factorization steps are extremely time-consuming when $n$ is large. To generate an $n\times n$ similarity matrix, we need to call the DTW algorithm $\mathcal{O}(n^2)$ times. 
Meanwhile, a naive implementation of eigen-decomposition takes $\mathcal{O}(n^3)$ time. We hereby introduce our Similarity PreservIng RepresentAtion Learning~(SPIRAL) framework that only calls DTW $\mathcal{O}(n\log n)$ times for similarity matrix construction, with an additional $\mathcal{O}(nd\log n)$ flops on learning $d$ features from the similarity matrix.

\subsection{The Low-rankness of the Similarity Matrix}
To significantly improve the efficiency of the first step, we make a key observation that the similarity matrix $\A$ can be well approximated by a low-rank matrix. In the following, we show that this observation is valid when the time series are generated from some distinguishable clusters.

\begin{assumption}\label{ass:clusters}
Suppose all the time series belong to $k$ clusters $C_1,C_2,\cdots, C_k$. For simplicity, let cluster $C_0$ to be the set of a unique length-one time series with entry 0. Define the cluster distance $d_{ab}$ between clusters $C_a$ and $C_b$ to be $d_{ab}=\min_{T_1\in C_a, T_2\in C_b}DTW(T_1,T_2), 0\leq a\neq b\leq k$. 
We consider the following conditions: 
\begin{enumerate}
\item Definition of clusters: 
   For each cluster $C_a$, we have $DTW(T_i,T_j)^2\leq \epsilon,\ \forall\ T_i, T_j\in C_a ,\ 1\leq a\leq k$;
\item Clusters are distinguishable: 
   $\epsilon\ll d_{ab}^2,\ \forall\  0\leq a\neq b\leq k$;
\item Proxy for triangle inequality: 
 For any two different clusters $C_a$ and $C_b$, $DTW(T_i, T_j)^2\leq d_{ab}^2+\mathcal{O}(\epsilon), \forall\  T_i\in C_a,\ T_j\in C_b,\ 0\leq a\neq b\leq k$;
\end{enumerate}
\end{assumption}

Under such assumptions, we have:
\begin{thm}
Let $T_1,T_2,\cdots T_n$ be $n$ time series generated from $k$ ($\ll n$) clusters satisfying Assumption \ref{ass:clusters}, 
then the generated similarity matrix $\A$ can be written as $\bL+\N$, where matrix $\bL$ has a small rank of at most $k(k-1)+2$, and $\N$ is a noise matrix satisfying $|\N_{ij}|\leq \mathcal{O}(\epsilon)$.
\end{thm}
We defer the proof to the appendix. Some alternative intuitions on the low-rankness of matrix 
$\A$ are: (i) time series similarity functions, including DTW, usually measure the level of co-movement between time series, which has shown to be dictated by only a small number of latent factors~\cite{stock2005implications}; and (ii) since the matrix $\A$ is a special case of Wigner random matrix, the gaps between its consecutive eigenvalues should not be small~\cite{marvcenko1967distribution}. This implies the low-rank property since most of its energy is concentrated in its top eigenvalues~\cite{erdHos2009local}. Our experiments on an extensive set of real-world time series datasets also verify the low-rankness of the similarity matrix $\A$.\footnote{We omit the details due to space limitation, but instead provide one example in Figure~\ref{fig:1}.}


\subsection{A Parameter-free Scalable Algorithm}
Given a low-rank similarity matrix, we are now able to significantly reduce the computational costs of both steps. 
According to the theory of matrix completion~\cite{sun2015guaranteed}, only $\mathcal{O}(n \log n)$ randomly sampled entries are needed to perfectly recover an $n\times n$ low-rank matrix. Thus, we don't need to compute all the pairwise DTW similarities. Instead, we randomly sample $\mathcal{O}(n \log n)$ pairs of time series, and then compute the DTW similarities within the selected pairs. This leads to a partially-observed similarity matrix $\tilde \A$ with $\mathcal{O}(n \log n)$ observed entries:
\begin{eqnarray}
\tilde \A_{ij}=\left\{
\begin{array}{ll}
\mathrm{S}(T_i, T_j) & \ \text{if\ }\ \Omega_{ij}=1 \\
\text{unobserved} &  \ \text{if }\ \Omega_{ij}=0,
\end{array}
\right.
\end{eqnarray}
where $\Omega\in\{0,1\}^{n}\times\{0,1\}^{n}$ is a binary matrix indicating the indices of sampled pairs. In this way, the running time of the first step is reduced by a significant factor of $\mathcal{O}(n/\log n)$. 
Since this factor scales almost linearly with $n$, we can significantly reduce the running time of generating the similarity matrix when $n$ is large.
As an example when $n=150,000$, it now only takes $194$ seconds to construct $\tilde{A}$ with [$20 n\log n$] observed entries, more than $500$ times faster than generating a full similarity matrix.

Given the partially-observed similarity matrix $\tilde \A$, our second step aims to learn a new feature representation matrix $\X$. 
To this end, we propose an efficient and parameter-free algorithm that can directly learn the feature matrix $\X\in \R^{n\times d}$ from minimizing the following optimization problem
\begin{eqnarray}
\min_{\X\in \R^{n\times d}}\ \ \ \|P_{\Omega}\ (\tilde\A\ -\X\X^\top)\|_F^2,
\label{eqn:fact}
\end{eqnarray}
where $P_{\Omega}: \R^{n\times n} \to \R^{n\times n}$ is a projection operator on $\Omega$. The objective function \eqref{eqn:fact} does not have a regularization term since it already bounds the Frobenius norm of $\X$. Despite its relatively simple formulation, solving problem \eqref{eqn:fact} is non-trivial since its objective function is non-convex. 
To address this issue, we propose a very efficient optimization algorithm that solves problem \eqref{eqn:fact} based on exact cyclic coordinate descent (CD). 
Our proposed coordinate descent algorithm has the following two advantages: (i) for each iteration, our CD algorithm directly updates each coordinate to the optimum. Thus, we do not need to select any hyper-parameter such as the learning rate; and (ii) by directly updating coordinates to the optimums using the most up-to-date information, our CD algorithm is efficient and converges at a very fast rate.

At each iteration of the exact cyclic CD method, all variables but one are fixed, and that variable is directly updated to its optimal value. 
To be precise, our algorithm consists of two loops that iterate over all the entries of $\X$ to update their values. The outer loop of the algorithm traverses through each column of $\X$ by assuming all the other columns
known and fixed. At the $i$-th iteration, it optimizes the $i$-th column $\X_{1:n,i}$ by minimizing the following subproblem
\begin{eqnarray}
\|\textbf{R}-P_{\Omega}(\X_{1:n,i}\X_{1:n,i}^\top)\|_F^2,
\end{eqnarray}
where $\textbf{R}$ is the residual matrix defined as $\textbf{R} = P_{\Omega}(\tilde \A-\sum_{j\neq i}\X_{1:n,j}\X_{1:n,j}^\top)$.
In the inner loop, the proposed algorithm iterates over each coordinate of the selected column and updates its value. Specifically, when updating the $j$-th entry $\X_{ji}$, we solve the following optimization problem:
\begin{eqnarray}
\begin{split}
& \min_{\X_{ji}}\ \!\!\|\textbf{R}-P_{\Omega}(\X_{1:n,i}\X_{1:n,i}^\top)\|_F^2\\
\!\!\!\Longleftrightarrow&\min_{\X_{ji}}\ \!\!\|\textbf{R}\|_F^2\!-\!2\langle \textbf{R}, P_{\Omega}(\X_{1:n,i}\X_{1:n,i}^\top)\rangle \!+\!\|P_{\Omega}(\X_{1:n,i}\X_{1:n,i}^\top)\|_F^2\\
\!\!\!\Longleftrightarrow&\min_{\X_{ji}}\ \X_{ji}^4+2(\!\!\!\!\sum_{k\in \Omega_j,\ k\neq j}\!\!\!\! \X_{ki}^2-\textbf{R}_{jj})\X_{ji}^2\\
&-4(\!\!\!\!\sum_{k\in \Omega_j,\ k\neq j}\!\!\!\! \X_{ki}\textbf{R}_{jk})\X_{ji}+C\\
\!\!\!\Longleftrightarrow& \min_{\X_{ji}}\ \psi(\X_{ji}):=\X_{ji}^4+2p\X_{ji}^2+4q\X_{ji}+C,\nonumber\label{eq:9}
\end{split}
\end{eqnarray}
where $\Omega_i,\ i=1,\!\cdots\!, n$ contains the indices of the observed entries in the $i$-th row of matrix $\tilde \A$, and $C$ is a constant. 
Algorithm \ref{alg:1} describes the detailed steps of the proposed exact cyclic CD algorithm that updates $\psi(X_{ji})$ to its minimum.

\begin{algorithm}[t]
    \caption{\ Efficient Exact Cyclic Coordinate Descent Algorithm for Solving the Optimization Problem \eqref{eqn:fact}}
    \begin{algorithmic}[1]
        \STATE {\bfseries Inputs:}
        \begin{itemize}
            \item $\tilde \A\in \R^{n\times n}$: partially-observed similarity matrix. $\Omega_i,\ i=1,\!\cdots\!, n$: indices of the observed entries in the $i$-th row of matrix $\tilde \A$. 
         $I$: \# iterations, $d$: \# features
        \end{itemize}
        \STATE {\bfseries Initializations:}
        \begin{itemize}
            \item $\X^{(0)}\leftarrow \mathbf{0_{n\times d}}$. $\textbf{R} \leftarrow P_{\Omega}(\tilde \A-\X^{(0)}\X^{(0)\top})$.
        \end{itemize}
        \FOR{$t=1,\cdots, I$}
        \STATE $\X^{(t)}\leftarrow \X^{(t-1)}$
        \FOR{$i=1,\cdots, d$}
        \STATE $\textbf{R}\leftarrow \textbf{R}+P_{\Omega}(\X^{(t)}_{1:n,i}\X_{1:n,i}^{(t)\top})$\label{step:add}
        \FOR{$j=1,\cdots n$}
        \STATE $p\leftarrow\sum_{k\in \Omega_j} \X_{ki}^{(t)2}-\X_{ji}^{(t)2}-\textbf{R}_{jj}$\\
        \STATE $q\leftarrow-\sum_{k\in \Omega_j}\X_{ki}^{(t)}\textbf{R}_{jk}+\X_{ji}^{(t)}\textbf{R}_{jj}$
        \STATE $\X_{ji}^{(t)}\leftarrow
        \argmin \{\X_{ji}^4+2p\X_{ji}^2+4q\X_{ji}\}$\label{step:root}
        \ENDFOR
        \STATE $\textbf{R}\leftarrow \textbf{R}-P_{\Omega}(\X^{(t)}_{1:n,i}\X_{1:n,i}^{(t)\top})$\label{step:subtract}
        \ENDFOR
        \ENDFOR
        \STATE {\bfseries Output:} $\X^{(I)}$
    \end{algorithmic}
    \label{alg:1}
\end{algorithm}

The proposed algorithm incurs a very low computational cost in each iteration. Lines $7$-$11$ of the algorithm can be computed in $\mathcal{O}(n\log n)$ operations. This is because the costs of computing $p$ and $q$ are only proportional to the cardinality of $\Omega_j$. Besides, the derivative $\nabla\psi(\X_{ji})$ is a third-degree polynomial, thus its roots can be computed in a closed form. By using Cardano's method~\cite{cardano1993ars}, the optimal solution of $\X_{ji}$ can be calculated in a constant time given the computed $p$ and $q$.
Likewise, lines $6$ and $12$ of the algorithm also take $\mathcal{O}(n\log n)$ time since matrix $\textbf{R}$ can be updated by only considering the observed entries. To sum up, the proposed algorithm has a per-iteration cost of $\mathcal{O}(d n\log n)$, 
which is significantly faster than direct factorization of the whole matrix that take at least $\mathcal{O}(dn^2)$ time in each iteration~\cite{DBLP:journals/corr/VandaeleGLZD15}.
The following theorem shows that Algorithm \ref{alg:1} guarantees to converge to a stationary point of~\eqref{eqn:fact}. We defer the proof to the appendix.
\begin{thm} \label{thm:CD}
Let $(\X^{(0)}, \X^{(1)}, \dots)$ be a sequence of iterates generated by the Algorithm~\ref{alg:1}. If it converges to a unique accumulation point, then the point is a stationary point of Problem \eqref{eqn:fact}.
\end{thm}

\begin{figure}[t]
    \centering
    \includegraphics[width=0.7\columnwidth]{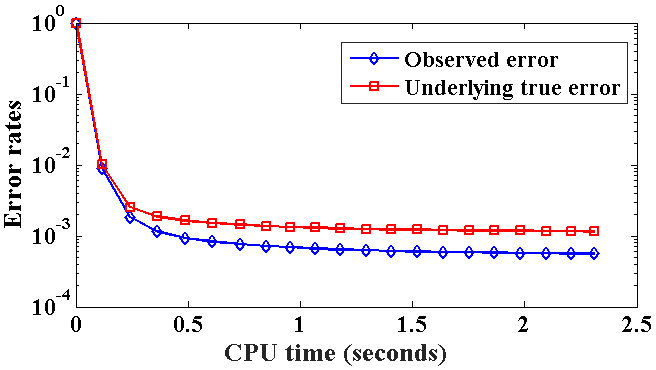}
    \caption{\small{Two error rates as a function of CPU time on UCR Non-Invasive Fetal ECG Thorax1 dataset}}\label{fig:1}
\end{figure}

In addition to a low per-iteration cost, the proposed algorithm yields a fast convergence. This is because our algorithm always uses the newest information to update variables and each variable is updated to the optimum in a single step. This is verified by a convergence test conducted on the UCR Non-Invasive Fetal ECG Thorax1 testbed~\cite{UCRArchive}. This testbed contains a total of $3,765$ time series with a length of $750$.
In this test, we generate a full similarity matrix $\A$ by computing the DTW similarities between all time series pairs, and then randomly sample $[20 n \log n]$ of its entries to generate a partially-observed matrix $\tilde \A$. We call the proposed algorithm to factorize matrix $\tilde \A$ by setting the dimensionality $d=15$. To measure the performance of the proposed method, we compute two error rates, i.e., the observed error $\|P_{\Omega} (\tilde\A\ -\X\X^\top)\|_F/\|P_{\Omega} (\tilde\A)\|_F$ and the underlying true error rate $\|\A\ -\X\X^\top\|_F/\|\A\|_F$, at each iteration. Figure~\ref{fig:1} shows how they converge as a function of time. This figure clearly demonstrates that the proposed exact cyclic CD algorithm converges very fast -- it only takes $1$ second and $8$ iterations to converge. Besides, the reconstruction accuracy is also very encouraging. The observed error and the underlying true error rates are close to each other and both of them are only about $0.1\%$. This result not only indicates that the inner products of the learned features well approximate the pairwise DTW similarities, but also verifies that we can learn accurate enough features by only computing a small portion of pairwise similarities. In addition, this test validates the low-rank assumption. It shows that a rank $30$ matrix can accurately approximate a $3,765 \times 3,765$ DTW similarity matrix. To further demonstrate the effectiveness and efficiency of our proposed algorithm, we compare our model with the naive gradient descent \cite{ge2016matrix}. For the 85 UCR time series datasets~\cite{UCRArchive}, our average running time is only 2.7 seconds, roughly 3 times more efficient than the gradient descent method \cite{ge2016matrix} with a learning rate $0.1$ to converge to a similar accuracy.

\vspace{-3pt}
\section{Experiments}
\vspace{-2pt}
{\bf Experimental setup:}
In this section, we thoroughly evaluate the proposed framework, i.e., Similarity PreservIng RepresentAtion Learning (SPIRAL for short), on the time series clustering task. We conduct extensive experiments on all the $85$ datasets in the UCR time series classification and clustering repository~\cite{UCRArchive} -- the largest public collection of class-labeled time-series datasets.

These datasets have widely varying characteristics: their training and testing set sizes vary from $16$ to $8,926$ and $20$ to $8,236$, respectively; the numbers of classes/clusters are between 2 and 60; and the lengths of the time series range from $24$ to $2,709$. Besides, this repository encompasses a wide range of domains such as medicine, engineering, astronomy, entomology, finance, and manufacture.

In our experiments, we set
$|\Omega|=[20n\log n]$, and \# features $d=15$. The convergence criteria is defined as the objective decreases to be less than 1e-5 in one iteration. 
Given the learned features, we then feed them into some static clustering models and compare them with some state-of-the-art time series clustering methods. To conduct fair comparisons, in all DTW related algorithms and all datasets, we set the DTW window size to be the best warping size reported in \cite{UCRArchive}.

To further verify that the proposed SPIRAL framework is flexible enough and works well on other similarity measures, we also conduct experiments that use move-split-merge (MSM) distance~\cite{stefan2013move} in our framework. We denote them as SPIRAL-DTW and SPIRAL-MSM, respectively.
All the results were averaged from 5 trials and obtained on a Linux server with an Intel Xeon $2.40$ GHz CPU and $256$ GB of main memory. Our source code and the detailed experimental results are publicly available. \footnote{\url{https://github.com/cecilialeiqi/SPIRAL} }



\begin{figure}[t]
    \centering
    \begin{tabular}{lr}
        \subfigure[\scriptsize{SPIRAL-DTW-$k$Means vs. $k$-Shape}]{\includegraphics[width=0.4\textwidth,height=0.3\textwidth]{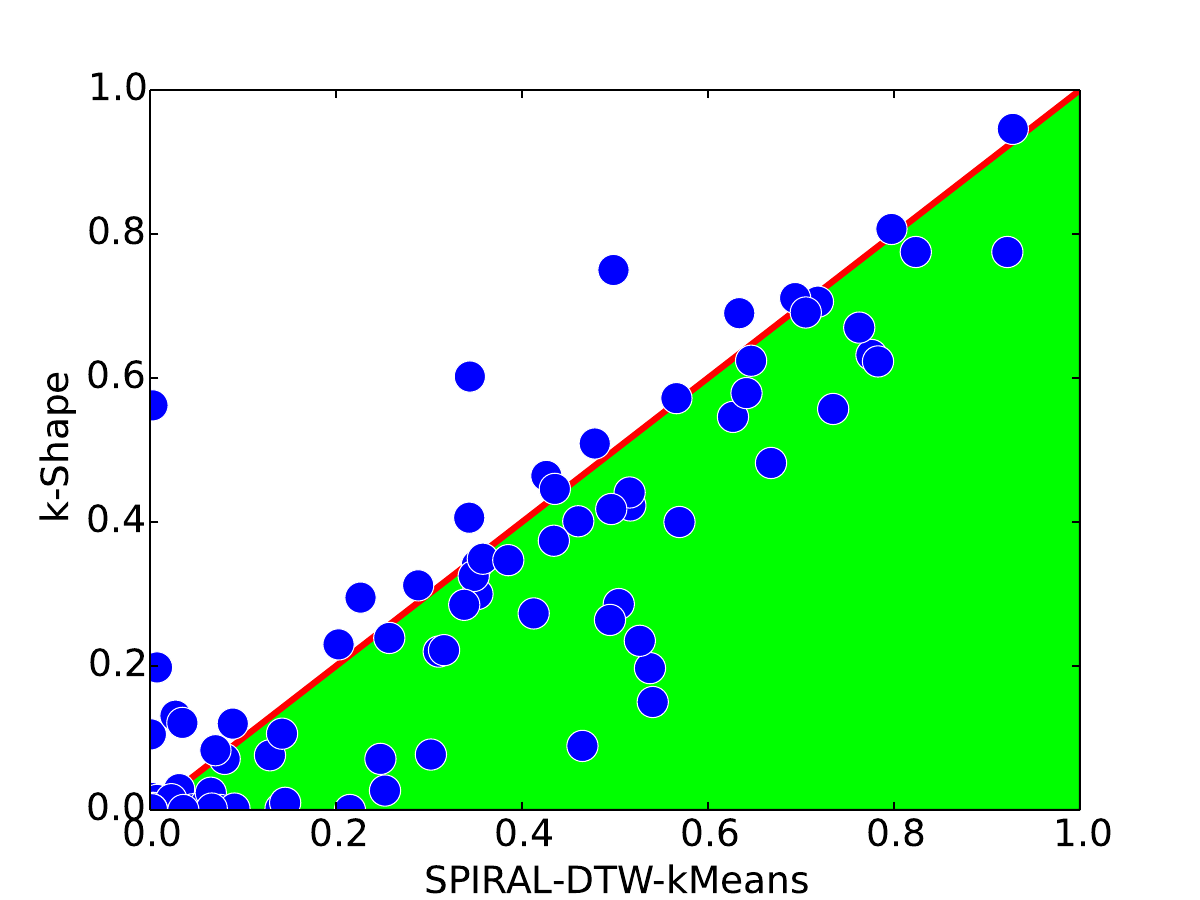}}
        \subfigure[\scriptsize{SPIRAL-DTW-$k$Means vs. CLDS}]{\includegraphics[width=0.4\textwidth,height=0.3\textwidth]{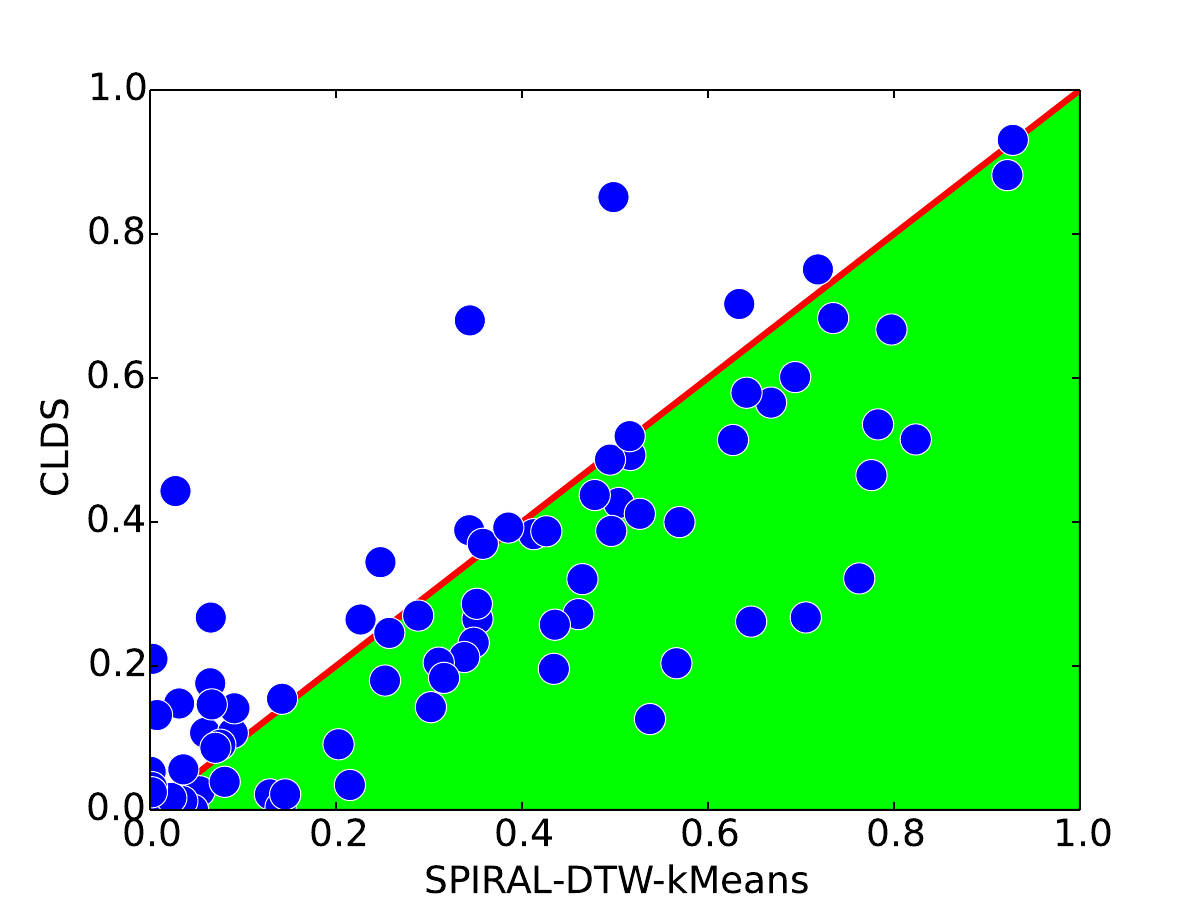}}\\
        \subfigure[\scriptsize{SPIRAL-MSM-$k$Means vs. Laplace-MSM-$k$Means}]{\includegraphics[width=0.4\textwidth,height=0.3\textwidth]{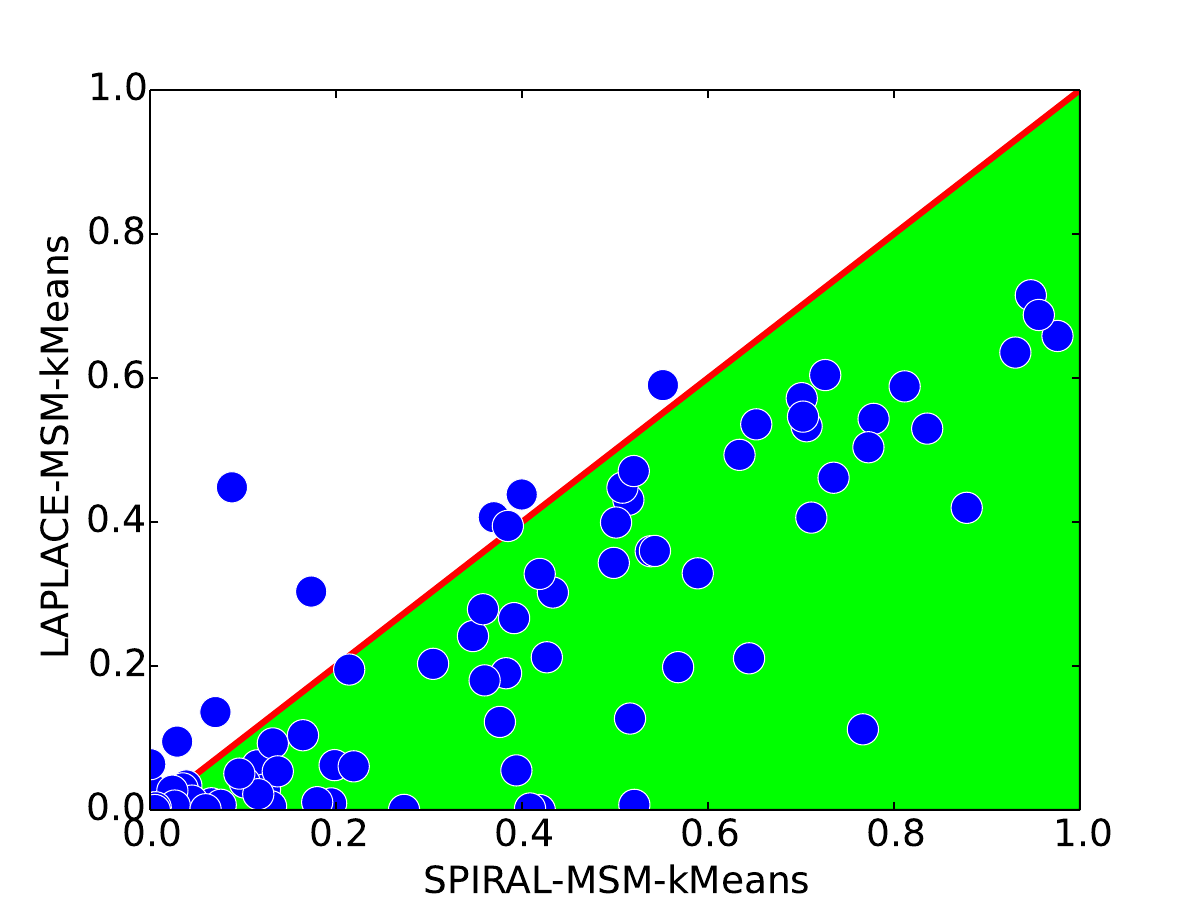}}
        \subfigure[\scriptsize{SPIRAL-MSM-$k$Means vs. $k$Medoids-MSM}]{\includegraphics[width=0.4\textwidth,height=0.3\textwidth]{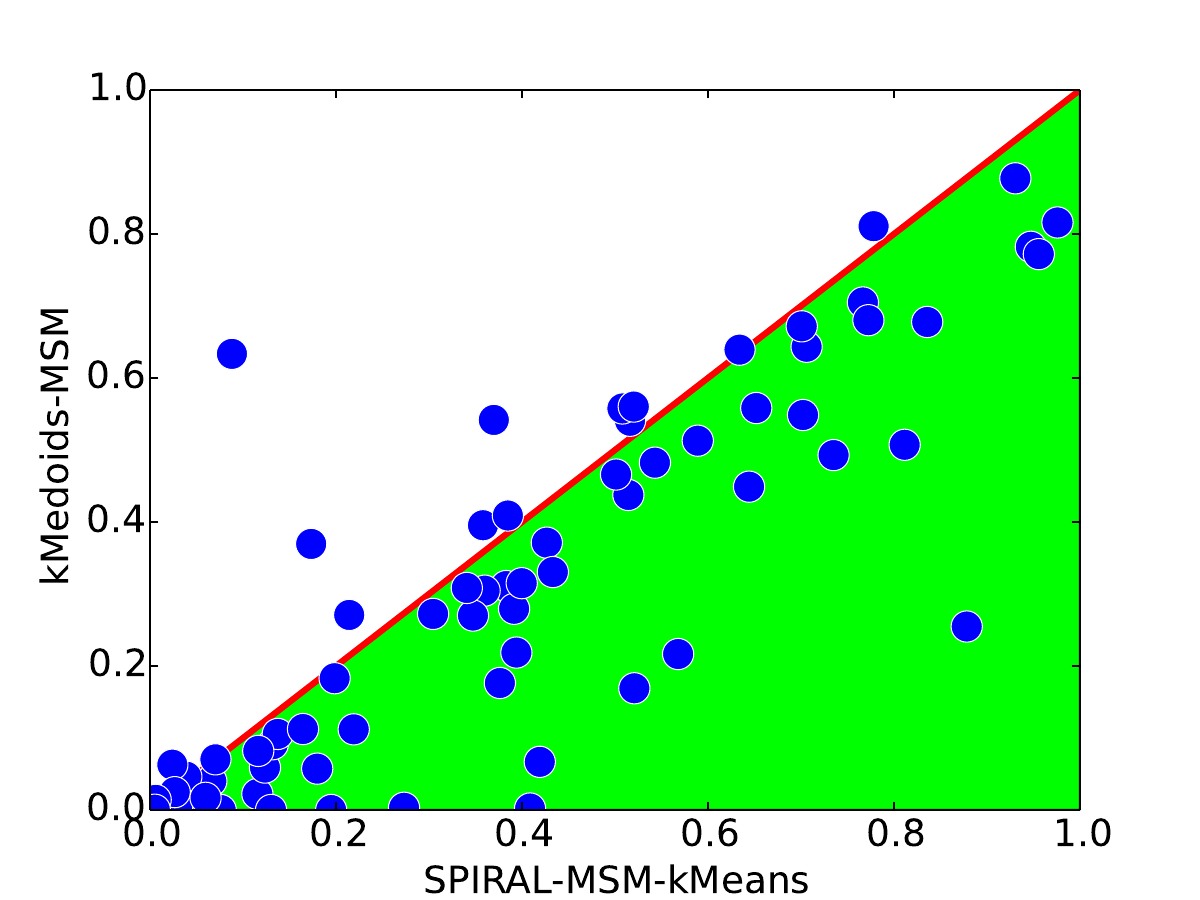}}        
    \end{tabular}
    \centering
    \caption{\emph{Comparison of our method with existing clustering algorithms over all the 85 UCR time series datasets.} 
    Top two: comparison between our method SPIRAL-DTW-$k$Means with two state-of-the-art methods $k$-Shape, and CLDS. Bottom two: comparisons between MSM based methods, our method SPIRAL-MSM-$k$Means with Laplace-MSM-$k$Means and $k$Medoids-MSM.  
    Circles below the diagonal indicate datasets over which our method yields better clustering performance in terms of NMI. 
} \label{fig2}
\end{figure}


Since data clustering is an unsupervised learning problem, we merge the training and testing sets of all the datasets. 
Given the features learned by the proposed framework SPIRAL with DTW similarity, we feed them into the $k$Means algorithm as our clustering method, denoted as SPIRAL-DTW-$k$Means. 

\begin{table*}[htb]
\centering
{\footnotesize
\begin{tabular}{c||c|c|c|c|c}
\toprule
method & {\bf SPIRAL-DTW-$k$Means} & {\bf SPIRAL-MSM-$k$Means} & Laplace-DTW-$k$Means & Laplace-MSM-$k$Means  & $k$-Shape\\
\hline
NMI	& {\bf 0.332} & {\bf 0.365} & 0.171	& 0.179	& 0.281\\
$<$(\%) & 59 (69.4\%) & N/A& 69 (81.2\%) & 67 (78.8\%)& 76 (89.4\%)\\
\hhline{======}
method & $k$Medoids-DTW & $k$Medoids-MSM & $k$Means-DTW & $k$Means-MSM & CLDS-$k$Means\\
\hline
NMI & 0.291	& 0.294	& 0.217& 0.235 & 0.285 \\
$<$(\%)& 65 (76.5\%) & 64 (75.3\%)  &72 (84.7\%) & 70 (82.4\%) & 63 (74.1\%)\\
\bottomrule
\end{tabular}
}
\caption{\emph{The overall clustering performance of all the proposed and baseline methods.} The notation $<$ denotes the number (percentage in parenthesis) of datasets over which the baseline methods perform worse than our method SPIRAL-MSM-$k$Means. 
}
\label{table:overall}
\end{table*}

{\bf Baseline methods:} 
We feed the features learned by our proposed framework into the $k$Means algorithm as our clustering method, denoted as \emph{SPIRAL-DTW-$k$Means}. To compare it with the existing time series representation methods, we respectively use the Laplacian transformation~\cite{hayashi2005embedding} and the complex-valued linear dynamical systems (CLDS)~\cite{li2011time} for extracting the same number of features as ours. We then feed the learned features into the $k$Means algorithm, forming two baselines \emph{Laplace-DTW-$k$Means} and \emph{CLDS-$k$Means}.  

Another baseline is the state-of-the-art time series clustering algorithm \emph{$k$-Shape}~\cite{paparrizos2015k}, which has been shown to outperform many state-of-the-art partitional, hierarchical, and spectral time series clustering approaches. Besides, we also compare our method with clustering algorithms \emph{$k$Means-DTW} and \emph{$k$Medoids-DTW} since our ideas are similar in some respects. $k$Means-DTW is a popular time series clustering algorithm that uses DTW algorithm to measure pairwise distances between data points. Although it looks similar to the idea of our SPIRAL-$k$Means that also utilizes the DTW and $k$Means algorithms, it is less desirable than SPIRAL-$k$Means mainly because: (i) $k$Means-DTW suffers from a very high computational cost since it needs to compute the pairwise DTW distances between all the time series and all the cluster centers at each iteration; and (ii) the DTW distance does not satisfy the triangle inequality, thus can make the cluster centers computed by averaging multiple time series drift out of the cluster~\cite{niennattrakul2007inaccuracies}.
By designing an efficient algorithm that only needs to call the DTW function $O(n \log n)$ times and by embedding time series data to the Euclidean space while preserving their original similarities, the proposed method SPIRAL successfully addresses both these issues. 





{\bf Experimental results:}
We use the normalized mutual information (NMI for short) to measure the coherence between the inferred clustering and the ground truth categorization. NMI scales from 0 to 1, and a higher NMI score implies a better partition. 
Figure~\ref{fig2} (a)(b) show that SPIRAL-DTW-$k$Means performs better in 62 (which is 72.9\%) and 52 (61.2\%) out of 85 datasets in comparison to $k$-Shape and CLDS, respectively. Though not plotted, it is also higher in 57 (67.1\%), 57 (67.1\%) and 66 (77.6\%) datasets compared with Laplace-DTW-$k$Means, $k$Medoids-DTW, and $k$Means-DTW. In all these comparisons, the statistical test demonstrates the superiority of SPIRAL-DTW-$k$Means. In addition to using DTW similarity in the SPIRAL framework, we also test our framework with another similarity measure move-split-merge (MSM), and denote this clustering method as SPIRAL-MSM-$k$Means. 
Figure~\ref{fig2} (c)(d) summarize the performance of all the MSM-based algorithms. 
The figures show that SPIRAL-MSM-$k$Means performs better in 69 (which is 81.2\%) and 64 (75.3\%) out of 85 datasets in comparison to Laplace-MSM-$k$Means and $k$Medoids-MSM, respectively. 
 Table \ref{table:overall} reports the average NMIs of all the algorithms, and the percentage of datasets over which the baseline methods perform worse than our algorithm SPIRAL-MSM-$k$Means. The table clearly shows that our proposed methods SPIRAL-MSM-$k$Means and SPIRAL-DTW-$k$Means yield the overall best performance. Besides, the MSM-based methods perform slighly better than the DTW-based methods, which is consistent with the observation in~\cite{DBLP:journals/datamine/BagnallLBLK17}. Moreover, the results also verify that the proposed method (\ref{eq:11}) is a better choice than the Laplacian method for converting DTW and MSM distances.
 

In addition to superior performance, our proposed framework has a significantly lower running time than all the baseline algorithms. 
For instance, clustering the ElectricDevice (ED) dataset with $16,637$ time series takes $k$-Shape, $k$Means-DTW and Laplace-DTW-$k$Means $20$ minutes, $169$ minutes and 5 hours, respectively. As a comparison, our clustering algorithms SPIRAL-DTW-$k$Means and SPIRAL-MSM-$k$Means only spend less than $2$ and $9$ minutes, respectively, to partition the ED dataset.

According to the extensive experimental results, we have the following observations:
\begin{itemize}
\vspace{-2pt}
    \item The proposed framework SPIRAL is flexible to multiple time series distance or similarity measures, including DTW and MSM. In more detail, the MSM-based method usually yields a better performance while the DTW-based method is more efficient. Both of them are orders of magnitude faster than the baseline algorithms.
\vspace{-2pt}
\item Our experimental results show that even simple clustering algorithms like $k$Means can yield strong performance on our learned representations. 
    By using some more advanced clustering algorithms, we may achieve an even better performance.
\vspace{-2pt}
\item The proposed framework learns a feature representation instead of directly developing a time series model. In this way, our method is more flexible and can exploit the strengths of different clustering algorithms.
\end{itemize} 
To sum up, the proposed framework is 
effective, efficient, and flexible. In addition to the superior performance, it is flexible in the choice of similarity measures and static learning algorithms. 
This enables the design of problem-specific algorithms to tackle various time series clustering problems.

\vspace{-3pt}
\section{Conclusions}
\vspace{-2pt}
In this paper, we show that the pairwise time series similarity matrix is close to low-rank, and propose a scalable representation learning framework that preserves pairwise similarities for time series clustering task. 
Our extensive empirical studies verify the effectiveness and efficiency of the proposed method.

\bibliographystyle{plain}
\bibliography{ref}
\appendix
\section{Proof of Theorem 2}
Using the definition of matrix $\A$ and assumptions $\mathnormal{A}$1-$\mathnormal{A}$3, we have
\begin{equation}
\A_{ij}=\left\{
\begin{array}{ll}
\frac{d_{a0}^2+d_{b0}^2-2d_{ab}^2}{2} +\Ocal(\epsilon) & \text{if }T_i\in C_a,\ T_j\in C_b,\ a\neq b \\
d_{a0}^2 +\Ocal(\epsilon) & \text{if }T_i,\ T_j\in C_a\nonumber
\end{array}\right.
\end{equation}
Let $N_{ij}=\Ocal(\epsilon)$, and
\begin{equation}
\bL_{ij}=\left\{
\begin{array}{ll}
\frac{d_{a0}^2+d_{b0}^2-2d_{ab}^2}{2} & \text{if }T_i\in C_a,\ T_j\in C_b,\ a\neq b \\
d_{a0}^2  & \text{if }T_i,\ T_j\in C_a,\nonumber
\end{array}\right.
\end{equation}
then $\A=\bL+\N$.

Let $I_a$ be the index of the time series in cluster $C_a$, $a=1,2,\cdots,k$, 
then the matrix $L$ could be divided into $k\times k$ blocks: $\bL_{I_a, I_b}, 1\leq a,b\leq k$, where each block has the same values:
\begin{equation}
\bL_{I_a,I_b}=\left\{
\begin{array}{ll}
\frac{d_{a0}^2+d_{b0}^2-2d_{ab}^2}{2} & \text{if }a\neq b\\
d_{a0}^2  & \text{if }a=b.\nonumber
\end{array}\right.
\end{equation}
Let $\x$ be a vector that satisfies $\x_{I_a}=\frac{d_{a0}^2}{2}$. For any index set $I$, let $\e^{I}$ be the indicator vector:
\begin{equation}
\e^{I}_i=\left\{
\begin{array}{ll}
1 & \text{if }i\in I\\
0 & \text{otherwise}\nonumber
\end{array} \right.
\end{equation}
Also, let $\mathbf{1}$ be the all 1 vector. Then we have:
\begin{equation}
\bL_{ab}=\x\mathbf{1}^\top+ \mathbf{1} \x^\top +\sum_{a\neq b} d_{ab}^2 \e^{I_a}(\e^{I_b})^\top. \nonumber
\end{equation}
In this sense, $\bL$ is the summation of $2+k(k-1)$ rank 1 matrices. Thus its rank is at most $2+k(k-1)$.

\section{Proof of Theorem 3}
Let $f(\X):=\|P_{\Omega}(\tilde{\A}-\X\X^\top)\|_F^2$, and $\bar{\X}$ be the unique accumulation point of the sequence $(\X^{(0)}, \X^{(1)}, \dots)$. With the coordinate descent algorithm, the generated sequence $f(\X^{(i)}), i=1,2,\cdots$ is monotonically non-increasing and bounded below. Note that only one variable has been updated between $\X^{(k+1)}$ and $\X^{(k)}$.

We prove the Theorem 2 by contradiction. Suppose $\bar{\X}$ is not a stationary point of problem (6), then there exists a pair $(i,j)$ satisfying
\[
f( \bar{\X} + \alpha \E^{ij} ) =  f(  \bar{\X} ) - \epsilon < f(  \bar{\X} ),
\]
where $\alpha \neq 0$, $\epsilon > 0$, and $\E^{ij}$ is the one-hot matrix with all zero entries except that the $(i,j)$'s entry equals to $1$.

Let $(\X^{(n_0)},\X^{(n_1)}, \dots)$ be a subsequence of $(\X^{(0)}, \X^{(1)}, \dots)$ and $n_k$ is the number of iterations that updates the entry $(i,j)$ for $k+1$ times.

Note that $f$ is continuous and $\X^{(n_k)}+\alpha \E^{ij}\rightarrow \bar{\X}+\alpha \E^{ij}$ when $k\to \infty$. There exists a sufficiently large $K$ so that for all $k > K$, we have
\begin{eqnarray*}
f( \X^{(n_k)}+\alpha \E^{ij}) &\leq& f(  \bar{\X} +\alpha \E^{ij}) +\frac{\epsilon}{2}\\
&=& f(\bar{\X})-\frac{\epsilon}{2}
\end{eqnarray*}
By flipping the sign of the above formula and adding $f( \X^{(n_k)})$ to both sides, we have
\[
 f( \X^{(n_k)} ) - f( \X^{(n_k)} + \alpha \E^{ij} )
 \; \geq \;
 f( \X^{(n_k)} ) - f(  \bar{\X} ) + \frac{\epsilon}{2} .
\]
By constructing the subsequence, the $(i,j)$th entry of $\X^{(n_k)}$ is updated earlier than the other entries to obtain $\X^{(n_{k+1})}$, which implies that
\[
f( \X^{(n_{k+1})} ) \leq f( \X^{(n_{k}+1)} ) \leq f( \X^{(n_k)} + \alpha \E^{ij} ).
\]
Hence we have
\begin{eqnarray*}
f( \X^{(n_k)} ) - f( \X^{(n_{k+1})} )
& \geq &
 f( \X^{(n_k)} ) - f( \X^{(n_k)} + \alpha \E^{ij} )
 \\
 & \geq & f( \X^{(n_k)} ) - f(  \bar{\X} ) + \frac{\epsilon}{2} \\
 & \geq & \frac{\epsilon}{2}.
\end{eqnarray*}
Since $f(  \bar{\X}) \leq f( \X^{(n_k)} )$, for all $k > K$ we have
\[
f( \X^{(n_{k+1})} ) \leq f( \X^{(n_k)} ) - \frac{\epsilon}{2},
\]
which leads to a contradiction since $f$ is bounded below.



\end{document}